\documentclass[10pt,twocolumn,letterpaper]{article}

\usepackage{cvpr}              % To produce the CAMERA-READY version
\definecolor{cvprblue}{rgb}{0.21,0.49,0.74}
\usepackage[pagebackref,breaklinks,colorlinks,allcolors=cvprblue]{hyperref}
\usepackage{multirow}

\def\paperID{*****} % *** Enter the Paper ID here
\def\confName{CVPR}
\def\confYear{2026}

\title{From Passive Execution to Active Exploration: Agentic Embodied Manipulation in Realistic Environments}

\author{
Shilin Ma, Chubin Zhang, Xulong Bai, Zifeng Gao, Shiyi Zhang, Yansong Tang$^{\dagger}$\\
Tsinghua Shenzhen International Graduate School, Tsinghua University
}

\begin{document}
\maketitle

\renewcommand{\thefootnote}{}
\footnotetext{$\dagger$ Corresponding author}

\begin{abstract}
Recent advances in agentic systems have substantially enhanced the long-horizon capability of embodied manipulation. However, many existing frameworks still follow a passive execution paradigm, which limits their applicability to real-world scenarios involving textual semantic cues, distractors, and initially invisible targets. To bridge this gap, we propose an agent-based active exploration framework that enables robots to dynamically interact with the environment rather than merely execute predefined instructions. Specifically, our framework consists of three collaborative modules: a planning module for high-level task reasoning, a perception module for visual scene understanding, and an execution module for low-level manipulation. This design allows the robot to actively acquire task-relevant information, adapt its behavior based on environmental feedback, and complete manipulation tasks under partial observability. Furthermore, we introduce a fine-grained perception–execution interleaving strategy, which tightly couples visual feedback with skill execution to improve exploration robustness. We evaluate our method on a realistic Find-and-Place task, demonstrating its effectiveness in challenging environments where target objects must be actively discovered before manipulation.
\end{abstract}    
\section{Introduction}
\label{sec:intro}

Recent progress in agentic systems has promoted the development of Vision-Language-Action (VLA) systems with stronger long-horizon reasoning and decision-making capabilities~\cite{li2026roboclaw,yang2025agentic,shi2026saivla,wang2026vla}. By integrating high-level task reasoning, agentic embodied systems extend robotic manipulation beyond isolated actions toward more complex sequential tasks. Instead of directly mapping an instruction to a short action sequence, these systems can decompose high-level goals into executable subtasks and maintain task progress across extended interactions. Such capabilities are essential for complex manipulation scenarios where robots must complete a sequence of dependent actions rather than a single short-horizon skill.

\begin{figure*}[tb]
  \vspace{-15pt}
  \centering
  \includegraphics[width=0.99\linewidth]{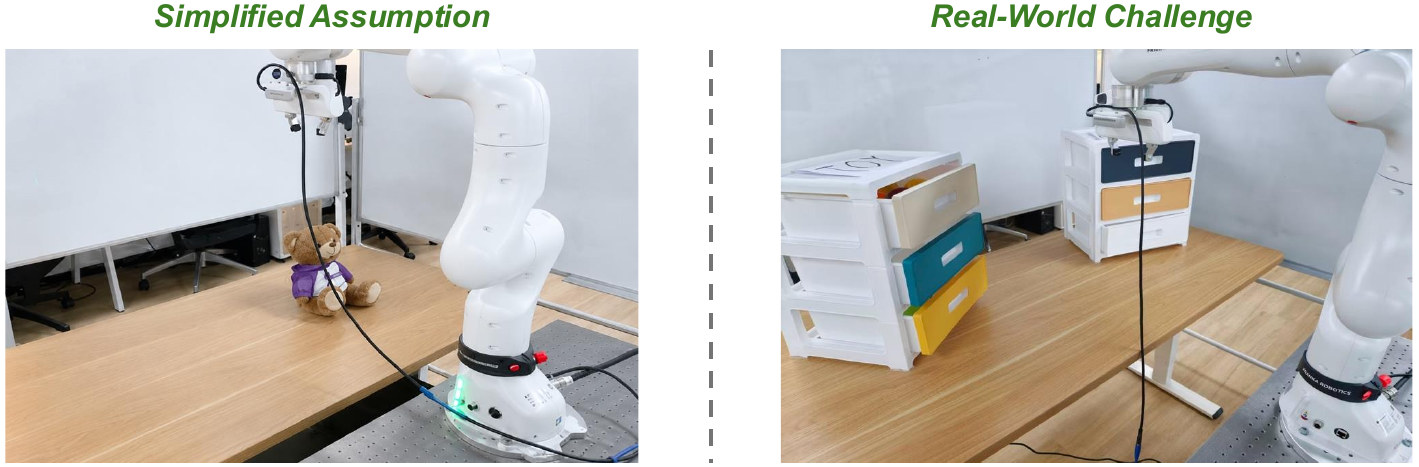}
  \caption{
  \textbf{From simplified assumptions to real-world active exploration.}
  Left: a simplified laboratory-style setting where the target object is visible and direct execution is sufficient. Right: a more realistic scenario where the target object is hidden in one layer of a cabinet and may be surrounded by distractors, requiring the robot to actively explore the environment. }
  \label{fig:comparison}
  \vspace{-15pt}
\end{figure*}

However, most existing embodied manipulation frameworks still follow a largely passive execution paradigm. Given a user instruction, the robot typically executes a predefined plan, policy, or skill sequence under \textbf{simplified assumptions}, such as visible task-relevant objects, uncluttered layouts, and sufficient visual information for direct execution. As shown in Fig.~\ref{fig:comparison}, these assumptions deviate from real-world scenarios, where target objects may be occluded, initially invisible, surrounded by distractors, or only identifiable through indirect semantic cues. While some planning-based methods~\cite{honerkamp2024language,menon2025open,schmalstieg2023learning,wang2025roboretriever} have considered such scenarios, they often rely on carefully engineered modules and less generalizable paradigms. This discrepancy leads to a significant gap between laboratory-style manipulation settings and practical real-world deployment.

To address this problem, we propose an agent-based \textbf{active exploration framework} for embodied manipulation. The central idea is to reformulate manipulation from passive instruction execution into a closed-loop process of perception, planning, and action. Specifically, our framework consists of three collaborative modules: a planning module that reasons about task progress and determines the next exploratory or manipulative step, a perception module that interprets visual observations and extracts task-relevant scene information, and an execution module that invokes low-level manipulation skills to interact with the environment. Through the coordination of these modules, the robot can explore partially observable scenes, update its scene understanding after interaction, and adapt its subsequent behavior according to environmental feedback. Furthermore, we introduce a fine-grained perception--execution interleaving strategy, which couples visual feedback with skill execution and enables more adaptive decision making throughout the manipulation process.

We evaluate the proposed framework on a realistic \textit{Find-and-Place} task, where the robot must locate a target object that may be initially invisible or hidden among distractors, and then place it at a specified target location. This setting reflects common challenges in real-world manipulation, including partial observability, semantic ambiguity, and the need for active interaction with the environment. Experimental results demonstrate that our framework improves the robustness and adaptability of embodied manipulation in challenging scenarios, validating the effectiveness of active exploration for narrowing the gap between simplified laboratory settings and real-world robotic deployment.

Our contributions can be summarized as follows:
\begin{itemize}
    \item We propose an agent-based active exploration framework for embodied manipulation, shifting robotic task execution from passive instruction following toward dynamic interaction with partially observable environments.

    \item We design a closed-loop architecture that integrates planning, perception, and execution, enabling robots to reason over visual feedback, invoke manipulation skills, and adaptively determine subsequent actions.

    \item We introduce a fine-grained perception--execution interleaving strategy and validate the proposed framework on a realistic \textit{Find-and-Place} task, demonstrating improved robustness under occlusion, distractors, and initially invisible targets.
\end{itemize}

\section{Method}
\label{sec:method}

We propose an agent-based active exploration framework for embodied manipulation, aiming to enable robots to complete tasks in partially observable and semantically complex environments. Given a high-level instruction, the robot is required not only to execute manipulation actions, but also to actively acquire missing information from the environment. This setting differs from conventional instruction-following manipulation, where the task-relevant objects are typically assumed to be visible and directly reachable. In contrast, our framework treats manipulation as a closed-loop decision-making process that continuously integrates perception, planning, and execution.

\begin{figure*}[t]
    \centering
    \includegraphics[width=0.95\linewidth]{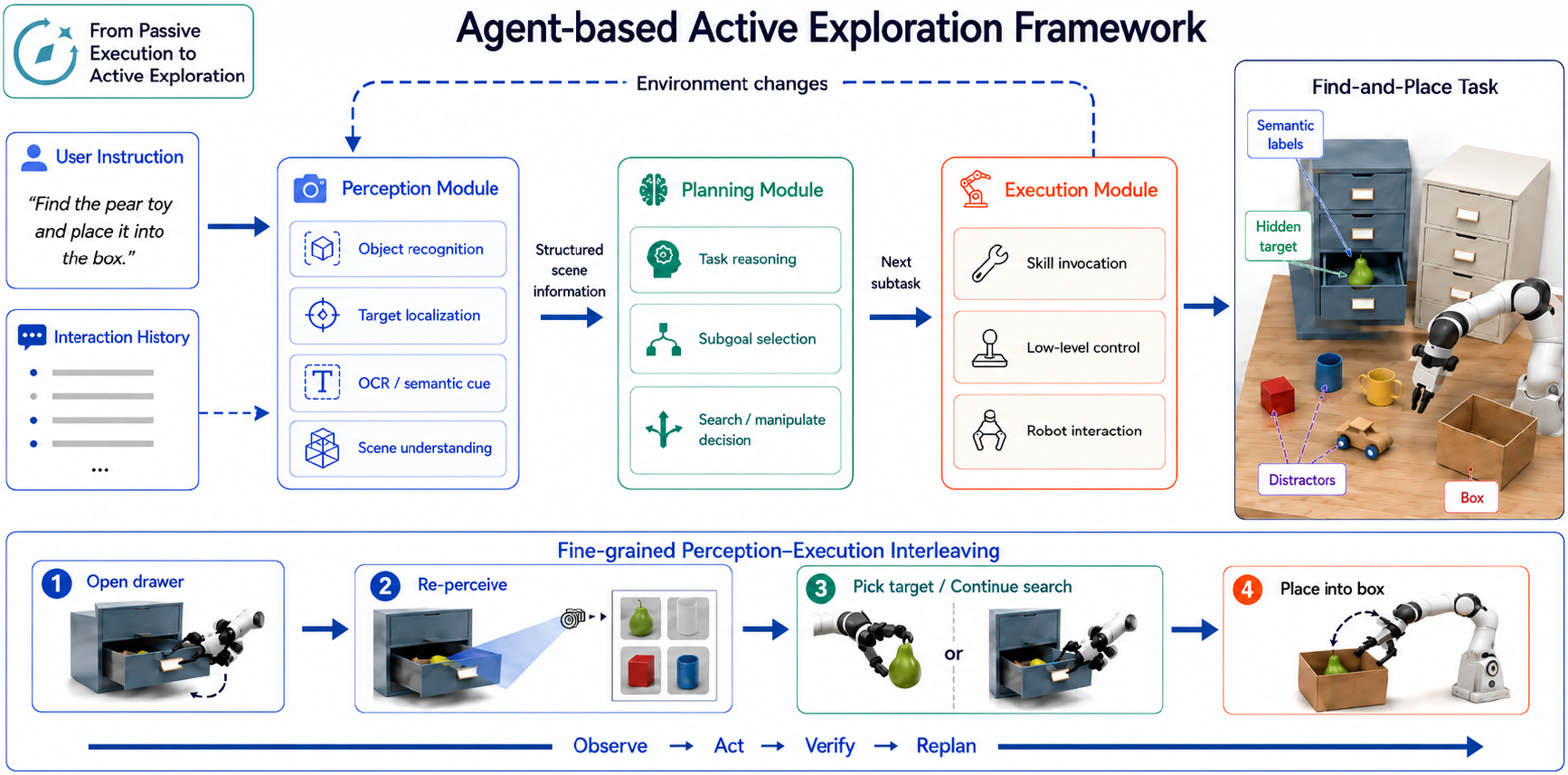}
    \caption{
    \textbf{Pipeline of our agent-based active exploration framework. }
    The system integrates perception, planning, and execution in a closed loop, enabling the robot to actively search hidden targets, update observations after each operation, and complete long-horizon Find-and-Place tasks through fine-grained perception–action interleaving.
    }
    \label{fig:pipeline}
    \vspace{-15pt}
\end{figure*}

\subsection{Overall Framework}
The framework consists of three collaborative modules: a planning module, a perception module, and an execution module. The planning module serves as the high-level decision maker. It takes as input the user instruction, the current visual observation, and the interaction history, and determines the next subtask to perform. Depending on the task progress, the planner may decide to search a specific region, inspect a semantic cue, open a container, retrieve an object, or place the object at a target location. In this way, the planning module converts an underspecified long-horizon instruction into a sequence of executable exploratory and manipulative steps.

The perception module provides structured scene understanding from visual observations and serves as the visual grounding component of the framework. It extracts task-relevant information through object recognition, target localization, optical character recognition (OCR), and scene-level semantic understanding. These perceptual outputs help identify visible objects, interpret textual or semantic cues, estimate relevant spatial regions, and determine whether the current scene contains sufficient information for task execution. By converting raw visual observations into structured environmental information, the perception module provides essential grounding for the planning module to make informed decisions.

The execution module performs the low-level manipulation process specified by the planning module. Given a planned subtask, it converts the high-level decision into executable robot actions and drives the robot to interact with the environment. Instead of requiring the agent to directly output continuous control commands, this module bridges task-level reasoning and physical execution. Such a separation improves execution stability in long-horizon tasks while allowing the planner to focus on semantic reasoning, task progress, and environmental feedback.

Through the cooperation of these three modules, the robot can actively interact with the environment and adapt its behavior according to feedback. At each decision step, the planner selects the next action based on the current scene understanding and task history; the execution module performs the corresponding manipulation skill; and the perception module updates the environment state after execution. This closed-loop formulation enables the robot to handle occlusion, distractors, and initially invisible targets, which are common in real-world manipulation scenarios but difficult for passive execution methods.

\subsection{Fine-grained Interleaving}

While the proposed modular framework provides the overall structure for active exploration, a key challenge lies in how to integrate perception into the manipulation process. A coarse strategy that performs perception only before or after the entire task is insufficient for partially observable environments, since each manipulation step may change the visible scene and reveal new task-relevant information. Therefore, effective active exploration requires perception and execution to be interleaved throughout the task, rather than two totally separate stages.

To this end, we introduce a perception–action interleaving strategy. Specifically, we decompose a long-horizon manipulation task into fine-grained operation units, such as opening a drawer, closing a drawer, and picking up an object. Each unit corresponds to an indivisible execution step that produces a potentially new environmental state. Instead of executing a complete action sequence based on the initial observation, the robot invokes the execution module for one atomic operation and calls the perception module to update its understanding of the environment.

This fine-grained interleaving enables the planner to make decisions based on the latest scene state. After each execution step, the perception module analyzes the updated observation and returns feedback about object visibility, scene changes, and task progress. The planning module then determines whether the current subgoal has been achieved, whether further exploration is required, or whether the task should proceed to the next operation. In this way, the robot does not passively follow a fixed plan, but observes, acts, verifies, and replans during execution.

This strategy is particularly important for tasks involving hidden or initially invisible objects. When the target object is absent from the current observation, the robot can actively explore the environment by selecting operation units that may reveal new information. For example, after opening a drawer, the robot immediately re-perceives the revealed space before deciding whether to pick an object, close the drawer, or continue searching elsewhere. Such step-by-step perception updates allow the system to progressively reduce uncertainty through interaction.

The proposed perception–action interleaving also improves robustness against distractors and semantic ambiguity. In cluttered scenes, multiple objects or regions may appear visually or semantically related to the instruction. By verifying the scene after each atomic operation, the framework can correct earlier assumptions and prevent errors from accumulating across long-horizon execution. As a result, the robot can perform manipulation as an active exploration process, improving its reliability in realistic environments with occlusion, distractors, and partial observability.

\section{Experiment}
\label{sec:exp}

\subsection{Real-World Task Setup}

\begin{figure}[t]
    \centering
    \includegraphics[width=0.95\linewidth]{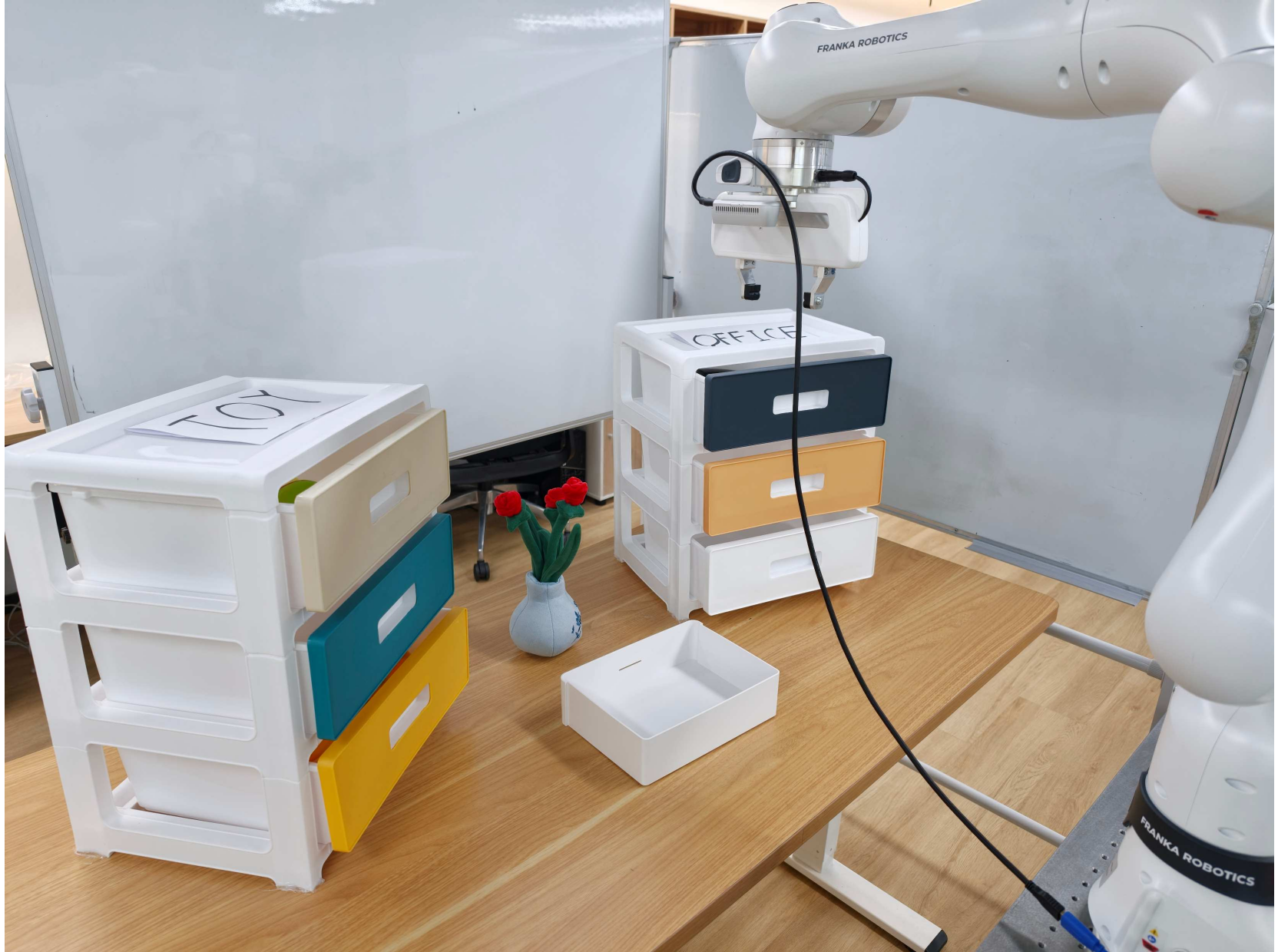}
    \caption{
    Real-world setup for the proposed \textit{Find-and-Place} task. 
    }
    \label{fig:environment}
    \vspace{-15pt}
\end{figure}

We evaluate the proposed framework on a real-world \textit{Find-and-Place} task designed to simulate practical manipulation scenarios with partial observability, semantic cues, and distractor objects. As shown in Fig.~\ref{fig:environment}, the environment contains two multi-drawer cabinets placed on a tabletop. Each cabinet is associated with a different object category and marked with a category label on its top surface. The target object is placed inside an unknown drawer and is not visible from the initial observation. In addition, each drawer may contain distractor objects, requiring the robot to actively search the environment rather than directly execute a predefined pick-and-place action.

For each trial, the user provides a natural language instruction in the form of:
\begin{quote}
``Could you help me find the \{target object\} and place it into the box?''
\end{quote}
Given this instruction, the system needs to infer the relevant cabinet from semantic category labels, sequentially inspect candidate drawers, identify the target object once it becomes visible, and finally place it into the box. This task setting evaluates whether the robot can perform active exploration under realistic conditions where the target is initially invisible and must be discovered through interaction.

\subsection{Implementation Details}

We deploy the proposed framework on a Franka robotic arm in the real world. The overall system is built upon OpenClaw, which serves as the agentic framework and planning module for high-level task reasoning. Given the user instruction and visual feedback, the planner determines the next exploration or manipulation step, such as selecting which cabinet or drawer to inspect and deciding whether the target object has been found.

The perception and execution modules are implemented locally. The perception module processes visual observations to support object recognition, target localization, optical character recognition, and scene-state understanding. The execution module converts the planned subtask into executable robot actions and controls the robot to interact with the environment. The entire system can run on a consumer-grade NVIDIA RTX 3090 GPU, demonstrating the practicality of the proposed framework without requiring large-scale inference or high-end data-center hardware.

\begin{figure}[t]
    \centering
    \includegraphics[width=0.95\linewidth]{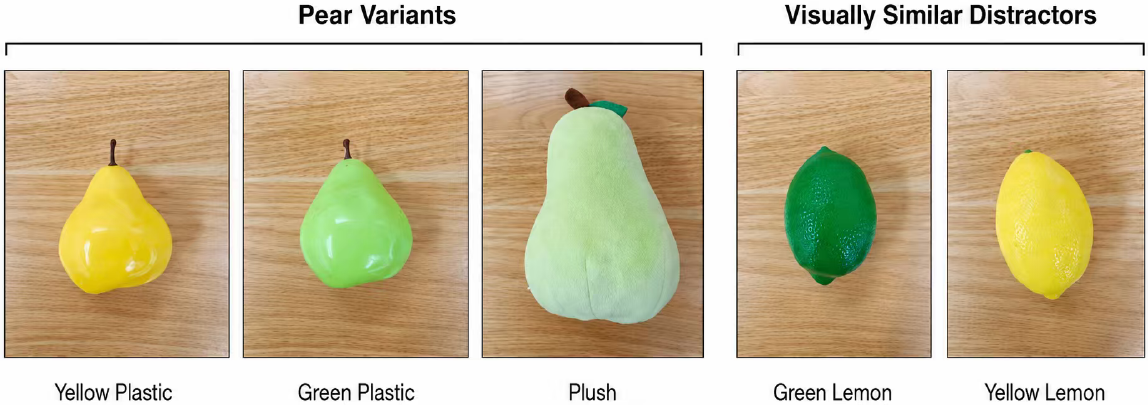}
    \caption{
    The evaluation includes variants with different appearances and visually similar distractors.
    }
    \label{fig:variant}
    \vspace{-15pt}
\end{figure}

\subsection{Evaluation Protocol}

We evaluate the proposed framework on the \textit{Find-and-Place} task with different target-object conditions. As shown in Fig.~\ref{fig:variant}, we evaluate one representative setting where the \textit{pear toy} is specified as the target object. Under this setting, we consider three target-object variants with different appearances, covering changes in color, size, material, and texture. These variants include two plastic pear toys with different colors and one plush pear toy with a different shape and texture. In addition, we introduce visually similar lemon-shaped distractors to evaluate the system's robustness under visual ambiguity.

A trial is considered successful only if the robot correctly identifies the relevant cabinet, searches the drawers until the target object is found, grasps the target object, and places it into the box. Failure cases include selecting the wrong cabinet, stopping before finding the target, grasping a distractor, or failing to complete the final placement.

\subsection{Results and Analysis}

Table~\ref{tab:find_and_place_results} reports the quantitative results on the \textit{Find-and-Place} task. We evaluate three target-object variants and compare two conditions: without visually similar distractors and with distractors. This setting tests whether the framework can remain reliable under both appearance variations and visual ambiguity.

\begin{table}[t]
    \centering
    \caption{Quantitative results on the \textit{Find-and-Place} task under different target-object variants and distractor conditions.}
    \label{tab:find_and_place_results}
    \begin{tabular}{lcc}
        \toprule
        \multicolumn{1}{c}{\multirow{2}{*}{\textbf{Setting}}} 
        & \multicolumn{2}{c}{\textbf{Success Rate (\%) $\uparrow$}} \\
        \cmidrule(lr){2-3}
        & \textbf{w/o Distractor} & \textbf{w/ Distractor} \\
        \midrule
        Target Object Variant 1 & 96.0 & 94.0 \\
        Target Object Variant 2 & 96.0 & 95.0 \\
        Target Object Variant 3 & 95.0 & 94.0 \\
        \midrule
        Overall & 95.7 & 94.3 \\
        \bottomrule
    \end{tabular}
    \vspace{-15pt}
\end{table}

The results show that our framework achieves consistently high success rates across all target-object variants, indicating strong generalization beyond a single fixed visual instance. Even with visually similar distractors, the success rate only slightly decreases from 95.7\% to 94.3\%, demonstrating that the proposed active exploration strategy can robustly identify, search for, and manipulate the intended target in cluttered and ambiguous environments.

\section{Conclusion}
\label{sec: conclusion}
In this work, we propose an agent-based active exploration framework for embodied manipulation in realistic settings. By integrating planning, perception, and execution into a closed-loop system, our framework enables the robot to actively acquire task-relevant information and adapt its behavior according to feedback. We further introduce a perception--action interleaving strategy, which updates scene understanding after each fine-grained operation to support more reliable long-horizon execution. Experiments on a real-world \textit{Find-and-Place} task demonstrate the effectiveness of our method in handling initially invisible targets, distractors, and category-guided search, highlighting the potential of shifting embodied manipulation from passive execution to active exploration.
{
    \small
    \bibliographystyle{ieeenat_fullname}
    \bibliography{main}
}

% WARNING: do not forget to delete the supplementary pages from your submission 
% \input{sec/X_suppl}

\end{document}